\documentclass[letterpaper]{article} % DO NOT CHANGE THIS
\usepackage[draft]{aaai25}  % DO NOT CHANGE THIS
\usepackage{times}  % DO NOT CHANGE THIS
\usepackage{helvet}  % DO NOT CHANGE THIS
\usepackage{courier}  % DO NOT CHANGE THIS
\usepackage[hyphens]{url}  % DO NOT CHANGE THIS
\usepackage{graphicx} % DO NOT CHANGE THIS
\usepackage{natbib}  % DO NOT CHANGE THIS AND DO NOT ADD ANY OPTIONS TO IT
\usepackage{caption} % DO NOT CHANGE THIS AND DO NOT ADD ANY OPTIONS TO IT
\usepackage{algorithm}
\usepackage{algorithmic}

\usepackage{newfloat}
\usepackage{listings}

\usepackage{arabtex}
\usepackage{utf8}
\setcode{utf8}

\usepackage{xcolor}
\usepackage{multirow}
\usepackage{stmaryrd}
\usepackage{hhline}
\usepackage{enumitem}

\DeclareCaptionStyle{ruled}{labelfont=normalfont,labelsep=colon,strut=off} % DO NOT CHANGE THIS
\floatstyle{ruled}
\newfloat{listing}{tb}{lst}{}
\floatname{listing}{Listing}
\def\linto{LinTO}
\def\lintoasr{\linto~ar-tn Kaldi ASR}
\def\noaug{\texttt{No-Aug}}
\def\novca{\texttt{No-VCA}}
\def\vca{\texttt{VCA}}

\title{\linto~Audio and Textual Datasets to \\ Train and Evaluate Automatic Speech Recognition in Tunisian Arabic Dialect}

\author {
    Hedi Naouara,
    Jean-Pierre Lorré,
    Jérôme Louradour
}
\affiliations {
    LINAGORA \\
    hnaouara@linagora.com, jplorre@linagora.com, jlouradour@linagora.com
}
\usepackage[hang,flushmargin]{footmisc}

\begin{document}

\maketitle 

\begin{abstract}
\begin{quote}

Developing Automatic Speech Recognition (ASR) systems for Tunisian Arabic Dialect is challenging due to the dialect's linguistic complexity and the scarcity of annotated speech datasets. To address these challenges, we propose the \linto~audio and textual datasets -- comprehensive resources that capture phonological and lexical features of Tunisian Arabic Dialect. These datasets include a variety of texts from numerous sources and real-world audio samples featuring diverse speakers and code-switching between Tunisian Arabic Dialect and English or French. By providing high-quality audio paired with precise transcriptions, the \linto~audio and textual datasets aim to provide qualitative material to build and benchmark ASR systems for the Tunisian Arabic Dialect.
%To enhance robustness, the dataset underwent noise reduction, and two augmentation techniques—voice conversion and standard perturbations, such as speed and volume adjustments—were applied to increase acoustic diversity.
% Its applications span ASR development, speech technology innovation, and language learning tools. The LinTO Audio Dataset plays a crucial role in advancing ASR research for Tunisian Arabic and could serve as a foundation for developing similar systems across other Arabic dialects.


\end{quote}
\end{abstract}


\section{Introduction}

In recent years, artificial intelligence has made significant advances in the fields of natural language processing (NLP) and automatic speech recognition (ASR).
Pioneering models such as wav2vec~\cite{baevski2020}, OpenAI Whisper~\cite{radford2022} and Meta Massively Multilingual Speech~\cite{tjandra2022massivelymultilingualasr70} boast support for dozens to hundreds of languages.
For example, despite massive amounts of training data (680,000 hours of transcribed speech for Whisper),
recent models fail to correctly transcribe with Arabic dialects.
For these reasons, we have collected data in the Tunisian Arabic dialect to provide high-quality resources for effective training and evaluation of ASR systems.
This paper presents our datasets and describes some preliminary results that show that they can be used to improve ASR sytems for Tunisian Arabic Dialect.

\subsection{Challenges and Existing Initiatives for Tunisian}

The linguistic evolution of the Tunisian Arabic dialect has been shaped by a complex historical background, influenced by various civilisations, including Amazigh (Berber)~\cite{tilmatine1999substrat}, Phoenician, Roman and Arab cultures, as well as the Ottoman Empire - now Turkey~\cite{ouerhani2009interference}, French colonialism and modern globalisation. These diverse influences have contributed to Tunisian lexical richness and distinct phonetic characteristics based on regional varieties~\cite{gibson1999dialect, ZRIBI2017147}, which pose significant challenges for the development of ASR systems. This task is further hindered by the limited availability of annotated speech datasets for Tunisian Arabic Dialect. %Furthermore, the limited availability of annotated speech datasets has posed additional obstacles to building robust ASR models specifically designed for this dialect.

Many Arabic dialects are underrepresented, including Tunisian Arabic. 
% This dialect presents unique challenges due to its phonetic complexities~\cite{gayraud2018}, the absence of standardized dictionaries~\cite{saidi2007typology}, a lack of formal writing rules, and frequent code-switching with French and English. 
This dialect presents unique challenges due to its phonetic complexity~\cite{gayraud2018}, lack of standardized dictionaries~\cite{saidi2007typology}, lack of formal writing rules, and frequent code-switching with French and English.
Although previous initiatives, such as TunSwitch~\cite{abdallah2023} and TARIC~\cite{mdhaffar-etal-2024-taric} have made progress, further efforts are needed to build a comprehensive dataset for Tunisian Arabic Dialect. 
%For instance, after examining both datasets, TARIC was found to be 
TARIC was collected entirely at a single train station, resulting in limited topics like greetings and ticket prices. TunSwitch is too small to train a model from scratch. And while the Massive Arabic Speech Dataset (MASC)~\cite{masc} contains %is also a dataset with 
1000 hours of speech collected from YouTube, it includes only 3 hours of Tunisian.
 
% tunisian dialec or Arabic

\subsection{Contribution}

To address these challenges, we present the \linto~datasets for Tunisian Arabic Dialect, comprising a diverse collection of text and audio contents.

The \linto~textual dataset in Tunisian includes content from various sources such as films and TV series, rap lyrics, documentaries, stories, and more.
This diverse collection was curated to support the training of language models for the Tunisian Arabic Dialect.
We also implemented tailored normalization to address the language's unique characteristics, such as standardizing the spelling of dialect-specific words, handling code-switching with French and English, and transliterating Arabizi. %% (a written form of Arabic using the Latin alphabet and numbers, commonly used in texting and online chats). 
These normalizations enhance the linguistic quality and coverage of the language model, especially when training data is limited.

The \linto~audio (raw and augmented) datasets in Tunisian include a wide range of recordings, such as music, documentaries, podcasts, TV shows, radio broadcasts, educational stories, and narratives of prophets, each accompanied by a transcript. To enhance dataset diversity, we employed data augmentation techniques on raw audio data. Specifically, Voice Conversion Augmentation was used to introduce more speaker variety into datasets with significant amounts of single-speaker recordings %large amounts of audio data for some speaker 
({\it e.g.} story telling).
% Additionally, noise reduction techniques were applied to enhance audio clarity and quality, addressing the challenges posed by the limited availability of annotated speech datasets. The cleaned audio can be subsequently augmented by adding noise, reverberation, and other effects.
Noise reduction techniques were applied to improve the clarity and quality of the audio, addressing the challenges posed by the limited availability of annotated speech datasets. Cleaned audio was then be enhanced by adding noise, reverberation and other effects.

Our datasets are designed to support speech recognition tasks for the Tunisian Arabic Dialect and feature  code-switching between Tunisian, English, and French. %It is 
They are meticulously organized into multiple configurations and splits to facilitate a variety of experimental setups, making them valuable resources for research and development in speech processing.
% The data has been carefully curated from a range of sources, including TunSwitch, Hugging Face, and other platforms.
We release three datasets under the CC BY 4.0 license on Hugging Face:
an audio dataset with transcribed speech in Tunisian\footnote{
%\textcolor{cyan}{\url{https://huggingface.co/datasets/linagora/linto-dataset-audio-ar-tn}}
\url{https://huggingface.co/datasets/linagora/linto-dataset-audio-ar-tn}
}, 
an augmentation of this dataset using Voice Conversion Augmentation\footnote{
%\textcolor{cyan}{\url{https://huggingface.co/datasets/linagora/linto-dataset-audio-ar-tn-augmented}}
\url{https://huggingface.co/datasets/linagora/linto-dataset-audio-ar-tn-augmented}
}
and a textual dataset gathering several sources of text in Tunisian Arabic Dialect\footnote{
%\textcolor{cyan}{\url{https://huggingface.co/datasets/linagora/linto-dataset-text-ar-tn}}
\url{https://huggingface.co/datasets/linagora/linto-dataset-text-ar-tn}
}.

In the following sections, we provide a detailed overview of the methodologies and procedures used to collect and process our Tunisian Arabic Dialect datasets. First, we discuss the compilation of textual corpora, highlighting the diversity of sources, the challenges % posed by code-switching between Tunisian Arabic, French, and English.
% Next, we describe the process of Arabizi transliteration and text standardization, which were essential to ensure linguistic consistency across the dataset.
and our solutions.
We then outline the strategies used for collecting transcribed audio recordings %, ensuring the inclusion of a wide range of speakers and contexts.
and describe data augmentation techniques.
Finally, we carry out some preliminary experiments to provide a first baseline ASR model for the Tunisian Arabic Dialect.

\section{Textual Corpus in Tunisian Arabic Dialect}

Language Models (LM) are a crucial element in many ASR systems.
They help determine the most likely spelling for a given sequence of phonemes, making their fine-tuning essential for optimal system performance.
Unlike Acoustic Models, which predict lattices of phoneme or character probabilities based on audio frames,
LMs do not require aligned audio data and can be trained solely on text, which is more cost-effective to collect.

\newcommand{\footnotesmall}[1]{}

The \linto~Tunisian text dataset was compiled from various sources,
including public datasets on Hugging Face:
\begin{itemize}[noitemsep, topsep=0pt]
    \item TuDiCoI:\footnotesmall{\url{https://huggingface.co/datasets/arbml/TuDiCoI}} A Tunisian dialogue dataset built by ARBML (An Arabic Researchers Community).
    \item Brahim Mohamed:\footnotesmall{\url{https://huggingface.co/datasets/medmabfc/Tunisien_Dialect_Summary-llama2-test26}} A Tunisian dialect summary dataset created to test Llama2.
\end{itemize}
several datasets from GitHub:
\begin{itemize}[noitemsep, topsep=0pt]
    \item T-HSAB:\footnotesmall{\url{https://github.com/Hala-Mulki/T-HSAB-A-Tunisian-Hate-Speech-and-Abusive-Dataset}} A Tunisian Hate Speech and Abusive Language dataset.
    \item TSAC:\footnotesmall{\url{https://github.com/fbougares/TSAC}} A Tunisian Sentiment Analysis Corpus.
    \item BARD:\footnotesmall{\url{https://github.com/4mekki4/arabic-nlp-da/tree/main/data}} A dataset of Arabic book reviews.
    \item Tunbert:\footnotesmall{\url{https://github.com/instadeepai/tunbert}} A dataset of daily Tunisian communications.
\end{itemize}
We include in addition
DrejjaToEnglish dataset from Kaggle\footnotesmall{\url{https://www.kaggle.com/datasets/khawlajlassi/drejja-to-english?resource=download}}
the TunSwitch dataset trancripts.
Finally, we crawled various blogs and websites:
\begin{itemize}[noitemsep, topsep=0pt]
    \item Stories: Chakhabit, HkayetErwi, TunHistoires, Lbachch\footnotesmall{\url{https://chakhabitt.blogspot.com/}, \url{https://hikayattunisien.blogspot.com/}}.
    \item Blog posts: TunHistoires, Lbachch\footnotesmall{\url{https://tunhistoires.blogspot.com/}, \url{https://lbachch.blogspot.com/}}.
    \item TV and film transcripts: ChroniqueChroniyet, KisatiAna\footnotesmall{\url{https://chroniquechroniyet.blogspot.com/}, \url{https://kisatiana.blogspot.com/}}.
    \item RAP lyrics: A Tunisian rap lyrics dataset\footnotesmall{\url{https://www.lyricstn.tn/search/label/RAP?m=1}}.
    \item Tweets: A collection of Tunisian Tweets.
\end{itemize}
The links corresponding to each source are provided in the dataset card of the \linto~dataset on Hugging Face.

This effort resulted in over 4.5 million lines of text and approximately 288,000 unique words.
The composition of the dataset is shown in Table~\ref{tab:sentence-distribution}.
Notably, the dataset includes sentences with code-switching between Tunisian and French or English,
as well as some sentences exclusively in French or English.
This inclusion enriches the vocabulary with terms sometimes used in spoken Tunisian.

\begin{table}[t]
\centering
\begin{tabular}{lrrr}
%\hline
&&&\textbf{\# unique}\\
\textbf{language(s)}        & \textbf{\# lines}  & \textbf{\# words}     &  \textbf{words}
\\
%\hline
\hline
Arabic
& 4\,269 k % 4\,268\,927
& 15\,964 k %\,299
& 255 k %\,486
\\ %\hline
French or English
& 207 k %\,454
& 594 k %\,167
& 28 k % 27\,841
\\ %\hline
code-switching
& 65 k % 64\,752
& 385 k %\,102
& 34 k % 33\,569
\\ %\hline
\hline
overall
& \bf 4\,541 k %\,133
& \bf 16\,944 k % 16\,943\,568
& \bf 289 k % 288\,677
%\\ \hline
\end{tabular}
\caption{
Composition of the \linto~Tunisian textual corpus.
%Language distribution within the text corpus used for training,
%depending on the type of sentence. % (Tunisian only, French or English only, and code-switching)
}
\label{tab:sentence-distribution}
\end{table}

% \subsection{Text Standardization}

The Tunisian Arabic Dialect lacks a standardized dictionary, leading to various spellings based on pronunciation~\cite{ouerhani2009interference}. For instance,
\RL{“هذا”} (meaning ``this'' or ``that'' and often pronounced as ``hatha'')
can appear as
\RL{“هاذا”}
or \RL{“هذى”}.
This variability complicates modeling and evaluation.
To address this issue, we created a normalization dictionary that maps more than 12,500 words to their standardized forms,
% This method enhances prediction accuracy by providing
to have
consistent representations of words.

% \subsection{Arabizi Transliteration}

Several materials  collected from social media and YouTube  include Arabizi, a writing style that mixes Latin letters and numbers to represent Arabic sounds.
% The term is a contraction of the Arabic word ``arabi'' (Arabic) and either ``inglizi'' (English) or ``easy''.
This challenges our models, which need text in Tunisian Arabic characters. For example, the Arabizi phrase ``9alou~y9awi~sa3dek~9alou~taw~taw~taw''
should be transliterated into ``\RL{قالوا يقوي سعدك قالوا تو تو تو}''.
%We initially struggled with automated translation tools for Arabizi.
We found  that models for arabizi transliteration such as~\cite{arabizi_lstm_2021}
are not suitable for the Tunisian Arabic Dialect, even though they perform well on Modern Standard Arabic.
To ensure accurate transcription, we semi-automatically transliterated Arabizi transcripts, by correcting a first automatic translation.
% \footnote{\url{https://www.tradonline.fr/blog/arabizi-langage-communication-ecrite/}}

\section{Transcribed Recordings of Spoken Tunisian}

The lack of annotated audio datasets is a significant challenge in developing an ASR system for dialects.
We focused on collecting clear and intelligible audio, representing diverse speakers and contexts.
In this section, we describe the data collected and the main pre-processing steps.

% DEPRECATED

\newcommand{\testnumbers}[2]{{~#1} & {~#2}}
\newcommand{\testnumbersbar}[2]{{~#1} & \multicolumn{1}{r|}{{~#2}}}
\newcommand{\detail}[1]{\textbf{\scriptsize{#1}}}
\def\hlinepartial{\cline{2-5}}
\def\undefined{\textcolor{red}{?}}
\newcommand{\factor}[1]{\scriptsize $\times {#1}$}

\begin{table}[t]
\centering
{\small
\begin{tabular}{@{}l@{}l@{}|r@{}r@{}r|@{}r@{}r}
&\multicolumn{1}{c}{}& \multicolumn{3}{c}{\textbf{TRAIN SET}}
& \multicolumn{2}{l}{\textbf{TEST SET }} % {TEST SET}
\\
\cline{3-7}
& 
& \textbf{\scriptsize audio }
& \textbf{\scriptsize VCA}
& \textbf{\scriptsize num.}
% & \multicolumn{2}{r}{\textbf{}} % {TEST SET}
& \testnumbersbar{\bf\scriptsize audio}{\bf\scriptsize num.}
\\
%\cline{6-7}
\textbf{\scriptsize source}
& \textbf{\scriptsize subset}
& \multicolumn{1}{r}{\textbf{\scriptsize dur.}}
& \textbf{\scriptsize aug.~}
& \textbf{\scriptsize words}
% & \multicolumn{2}{l|}{\scriptsize \textbf{dur. / words}}
% & \scriptsize \textbf{dur.} & \multicolumn{1}{l|}{\scriptsize \textbf{words}}
& \testnumbersbar{\scriptsize \textbf{dur.}}{\scriptsize \textbf{words}}
\\
%\hline%\hline
\hhline{==:===:==}
\multirow[t]{2}{*}{\textbf{TunSwitch~}}
& \textbf{CS} % {\tiny (Code-Switching)}
& 10H01 & \factor{7.0} & 75 k & \testnumbers{27m}{4 k}
\\
%\multirow{2}{*}{\textbf{TunSwitch}}
& \textbf{TO} % {\tiny (Tunisian Only)}
& 3H20 & \factor{6.7} & 18 k & \testnumbers{28m}{3 k}
\\
\hlinepartial
\textbf{Hugging}
& \detail{Ameni Kh}
& 4H05 & \factor{1} & 32 k & \testnumbers{3m}{{\scriptsize 0.5} k}
\\
\multirow[t]{2}{*}{\textbf{\quad Face}}
& \detail{Arbi Houssem}
& 3H50 & \factor{1} & 33 k & 
\\
& \detail{MA Konyali}
& 3H27 & \factor{1} & 20 k &
\\
\hlinepartial
\multirow[t]{3}{*}{\textbf{YouTube}}

& \detail{story telling} % AbdelAzizErwi + HkeytetTounsiaMensia + LobnaMajjedi
& 27H17 & \factor{6.2} & 148 k & % (122h 51m 1s + 	12h 13m 29s	 + 6h 41m 38s + 27h 16m ) / 27h 16m
\\
& \detail{theme channels} % BayariBilionaire + HamzaBaloumiElMohakek + QLM
& 20H12 & \factor{1.3} & 113 k &
\\
& \detail{TV \& radio} % TV + DiwanFM + MohamedKhammessi
& 7H48 & \factor{8.0} & 48 k & % 
\\
& \detail{misc. crawled}
& 4H08 & \factor{8.8} & 19 k & \testnumbers{19m}{1 k} % 4H10
\\
& \detail{shorts}
& 3H47 & \factor{8.0} & 28 k &
\\
& \detail{MASC}
& 2H53 & \factor{7.9} & 12 k &
\\
\hlinepartial
\multirow[t]{2}{*}{\textbf{websites}}
& \textbf{\scriptsize OneStory}
& 1H33 & \factor{7.1} & 12 k & \testnumbers{3m}{1 k}
\\
& \textbf{\scriptsize Appr.LeTunisien}
& 0H38 & \factor{5.2} & 1 k & \testnumbers{3m}{{\scriptsize 0.15} k}
\\
% \cline{1-5}
% \hline
% \hhline{:--:===:--.}
%\hhline{~~:===:--}
\hhline{==:===:==}
% \linto~ar-tn~~
\multicolumn{2}{r|}{\it overall}
& \bf 92H56 
& %$\shortrightarrow$ 
\bf ~ 466H 
& \bf ~ 560 k
& \testnumbersbar{\bf 1H20}{\bf 10 k}
\\
% \hline
% \hline
\multicolumn{2}{r|}{\textbf{TARIC} {\scriptsize(not in released data)} }
& 7H25 
& 52H 
& 57 k 
& \testnumbersbar{0H50}{7 k}
\\
\cline{3-7}
\end{tabular}
}
\caption{
Composition of the \linto~Tunisian audio dataset and TARIC.
Audio durations are indicated with/without VC Augmentation.
Number of transcribed words are also given.
%Durations indicates transcribed speech (without silence nor music segments).
}
\label{tab:audio-data-composition}
\end{table}

\subsection{Dataset Composition}

Capturing diverse contexts in the training data is essential for effective speech modeling. 
% Many existing datasets lack variety.
To enhance diversity, we gathered several existing datasets
and collected  dozens of additional hours of data by crawling the web. 
The \linto~audio dataset, whose composition is detailed in Table~\ref{tab:audio-data-composition},
includes:
% \\ \textcolor{red}{TODO (below): describe the type of data: read/spontaneous/conversational speech, specify when the lexical fields is limited (train station, simple vocabulary to learn the language...)}
\begin{itemize}[noitemsep, topsep=0pt]
% ~\cite{abdallah2023}
    \item the transcribed part of TunSwitch~\cite{abdallah2023}, which can be divided into two parts: 75\% with code-switching (CS) and 25\% with Tunisian Arabic only (TO)
    \item several datasets available in Hugging Face:
    \begin{itemize}[noitemsep, topsep=0pt]
        \item Arbi Houssem:\footnote{\tiny\url{https://huggingface.co/datasets/Arbi-Houssem/Tunisian_dataset_STT-TTS15s_filtred1.0}} TV shows
        \item Ameni Kh:\footnote{\tiny previously under \url{https://huggingface.co/datasets/amenIKh/dataset1}} TV and radio content
        % a portion of the TunSwitch Weakly audios, including content from radio and TV shows, was manually annotated.
        \item MA Konyali:\footnote{\tiny previously under \url{.../medaminekonyali/Value-Wav2Vec-tunisian-Darja-Augmented}}  isolated words with augmentation
    \end{itemize}
    \item the Tunisian subset of MASC~\cite{masc}, comprising YouTube audios with provided transcripts. We re-extracted audio and transcripts from MASC URLs using our YouTube crawling pipeline.
    
    % the Tunisian part from MASC~\cite{masc}, which are audios from YouTube with cautions used as ground-truth transcripts. We decided to re-extract ourselves audio and transcripts from the URLs found in MASC, based on the pipeline we developed to crawl transcribed audio from YouTube.
    %________
\end{itemize}
Our experiments include TARIC~\cite{mdhaffar-etal-2024-taric} 
but we do not re-distribute it within the \linto~audio dataset
due to its non-commercial license.

We also crawled Tunisian educational platforms such as ApprendreLeTunisien\footnote{\url{https://www.apprendreletunisien.com/}} and OneStory Media\footnote{\url{https://www.onestory-media.org/}},
and YouTube videos with curated captions taken as transcripts.
All audio recordings are sampled at 16~kHz.
The resulting data fall into the following categories:
\begin{itemize}[noitemsep, topsep=0pt]
\renewcommand{\labelitemi}{--}
\item {\bf story telling}:
Abdel Aziz Erwi,
Hkeytet Tounsia Mensia,
Lobna Majjedi
\item {\bf theme channels}:
Bayari Bilionaire (soccer),
Hamza Baloumi El Mohakek (crime),
QLM media (history)
\item {\bf TV and radio}:
Carthage plus and Telvza TV (TV)
Diwan FM (radio),
Mohammed Khammesi (podcast)
\item {\bf short videos}: Short videos of  radio shows, announcements and some manually selected jokes
\item {\bf misc. crawled videos}: This data was manually collected by reviewing content to select annotated audio from diverse channels. We faced challenges because YouTube does not filter audios by specific dialects, so we manually identified Tunisian audios from those tagged as Arabic. 

% for instance, the wrong annotation, native Arabic annotation, Arabizi text and more      

% \textcolor{red}{TODO
%     Explain the difficulty to filter data in Tunisian (no dialect classifier that includes Tunisian, "Tunisian" is not a tag in YouTube caption, all dialects are under "Arabic").
%     Explain our approach : selection of YouTube channels.
%     It would have been great to generate search queries in Tunisian and try them on YouTube, but I think we haven't done that.
% }

\end{itemize}

% This approach significantly enriched our dataset in terms of contextual and vocal diversity.

% \textcolor{red}{TODO 
% put a pie with the composition of the dataset,
% and be clear about total duration of the dataset (before VCA)
% }
% \begin{figure}[ht]
%     \begin{minipage}{0.48\textwidth}
%         \centering
%         \includegraphics[width=\textwidth]{figs/pie.png}
%         \caption{
%   Distribution of audio durations across \linto~Dataset 
%         }
%         \label{fig:data_pie}
%     \end{minipage}
% \end{figure}

\subsection{Transcript Curation}

Combining existing datasets with online sources highlighted challenges like incorrect audio annotations. Accurate annotations are critical for training effective Speech-to-Text models and must align with our standardization dictionary.

% We addressed these issues using specialized tools. For unannotated or incorrectly annotated audio, we employed Audacity for analysis and Praat for correcting and segmenting the audio, improving dataset reliability.

Over 85\% of our YouTube dataset was transcribed using YouTube's Auto-Transcribe.
% \textcolor{red}{Is "Auto-Transcriber" the right term? Isn't it Auto-Transcribe ?}
While the  results were generally acceptable, some issues arose, such as word truncation (e.g., \RL{“تع”} and \RL{“نع”} for \RL{“متاع”}) and the substitution of incorrect words (e.g., \RL{“معنتها”} was frequently transcribed as \RL{“مع”}). To improve transcription quality, we referred to a standardized dictionary that included both truncated and complete forms of words. For more complex cases, we manually verified the transcripts by listening to the corresponding audio segments. %, ensuring greater accuracy.

\subsection{Segmentation of Long Audio Segments}

ASR systems are usually trained on 15 to 30 second audio segments.
We thus segmented longer audio files into chunks not exceeding 30 seconds.
As described in~\cite{zhu2021}, we implemented a forced alignment to accurately align audio segments with their respective transcripts,
based on a wav2vec ASR model trained on Modern Standard Arabic \cite{grosman2021xlsr53-large-arabic}.
%, we performed alignment without requiring precise word-level predictions; the model only needed to recognize Arabic phonemes and perform segmentation for accurate audio-text alignment.
This alignment procedure was also used to correct misalignments in YouTube captions. %, ensuring precise segment alignment across the dataset.

% To optimize model training, we divided longer audio segments into smaller chunks, especially those exceeding 30 seconds, to improve processing efficiency. We implemented a script that leverages a pre-trained wav2vec ASR model for Modern Standard Arabic. This model’s role is to recognize and segment Arabic words without necessarily requiring precise word predictions, an effective approach for low-resource languages and dialects. Additionally, the script corrected mismatched annotations by aligning audio segments with their corresponding transcripts, ensuring better data quality and consistency~\cite{article}.

% the original duration of the MASC dataset was 2 hours and 53 minutes, and the crawled data was 4 hours and 10 minutes. After applying noise reduction tools, we doubled the data size for both datasets, as shown in Table~\ref{tab:audio-data-composition}.
% This approach allows us to go for reverberation 

\section{Augmentation of Audio Data}

%\subsection{Classical Augmentation}

% \textcolor{red}{TODO 
% clarify that these classical augmentation technics are done just before training, are fast to do. Augmented data are not included in the \linto~audio corpus, contrarily to VC Augmented data.}

Several data augmentation techniques are used to improve the robustness of acoustic modeling. These methods expand the training dataset without the need for real speech data,
helping the model handle variations in speakers, recording conditions and background noises.
Standard techniques include: speed change,
% One widely used technique is speed perturbation, which adjusts the playback speed of the audio by either increasing or decreasing it. This helps the model better adapt to different speaking rates.
volume change,
% Another technique is volume perturbation, which changes the volume to simulate different recording conditions.
simulated reverberation and background effects.
% Lastly, reverberation adds simulated background effects, like room echoes, to help the model perform well in diverse acoustic settings.
% These techniques have proven effective in enhancing ASR systems. In our case, however, we used only two augmentation types: speed perturbation and volume perturbation. Since our dataset already contains many noisy audios, adding reverberation would risk further degrading the data quality.
These standards or classical data augmentation techniques are generally applied just before the training phase and are cheap to compute. Contrary to the augmented data described below,
such ``classical'' augmentation is not included in the \linto~audio raw corpus.

\begin{table*}[!ht]
\resizebox{\textwidth}{!}{ % Resize the table to fit within text width
%\begin{tabular}{l|r|r|r|r|r|r}
\begin{tabular}{ll@{}rrrrrr}
\multirow{2}{*}{\textbf{ASR}} && \multirow{2}{*}{\textbf{YouTube}} & \multicolumn{2}{c}{\textbf{TunSwitch}}  & \textbf{Apprendre} & \multirow{2}{*}{\textbf{TARIC}} & \multirow{2}{*}{\textbf{OneStory}} \\
&&  & \textbf{CS} & \textbf{TO} & \textbf{Le Tunisien} &  &  \\
%& \textbf{WER | CER} & \textbf{WER | CER} & \textbf{WER | CER} & \textbf{WER | CER} & \textbf{WER | CER} & \textbf{WER | CER} \\
 %   & \textbf{WER} 
 %   & \textbf{WER} 
 %   & \textbf{WER} 
 %   & \textbf{WER} 
 %   & \textbf{WER} 
 %   & \textbf{WER}
 % \\
\hline
OpenAI's Whisper & \texttt{large v3}
   & 82.0% \% | 44.72% \% 
   & 93.8% \% | 63.12% \% 
   & 60.4% \% | 22.89% \% 
   & 66.7% \% | 30.63% \% 
   & 92.6% \% | 56.17% \% 
   & 49.8% \% | 14.78% \%
 \\
& \texttt{large v3 turbo}
   & 89.6% \% | 52.88% \% 
   & 117.9% \% | 86.52% \% 
   & 62.9% \% | 24.07% \% 
   & 74.8% \% | 31.59% \% 
   & 110.1% \% | 65.39% \% 
   & 53.1% \% | 15.11% \%
 \\
& \texttt{large v2}
   & 78.7% \% | 47.58% \% 
   & 101.2% \% | 69.87% \% 
   & 62.3% \% | 25.18% \% 
   & 66.7% \% | 30.22% \% 
   & 102.8% \% | 66.26% \% 
   & 54.1% \% | 17.07% \%
 \\
% Whisper -- medium 
%    & 100.9% \% | 57.00% \% 
%    & 116.7% \% | 75.73% \% 
%    & 74.4% \% | 35.28% \% 
%    & 74.2% \% | 32.28% \% 
%    & 104.4% \% | 63.62% \% 
%    & 56.8% \% | 17.42% \%
%  \\
% Whisper -- small 
%    & 113.1% \% | 78.35% \% 
%    & 129.5% \% | 84.81% \% 
%    & 76.8% \% | 32.35% \% 
%    & 81.1% \% | 31.73% \% 
%    & 118.8% \% | 75.04% \% 
%    & 66.7% \% | 20.62% \%
%  \\
 \hline
\lintoasr & \noaug % - baseline
   &  47.8% \% | 36.62% \% 
   &  58.5% \% | 60.32% \% 
   &  35.5% \% | 32.80% \% 
   &  28.3% \% | 28.16% \% 
   &  25.7% \% | 22.60% \% 
   &  26.9% \% | 43.91% \%
 \\
   & \novca % - baseline
   & \bf 34.4% \% | 36.62% \% 
   &  25.1% \% | 60.32% \% 
   & \bf 20.6% \% | 32.80% \% 
   &  26.4% \% | 28.16% \% 
   & \bf 13.9% \% | 22.60% \% 
   &  4.8% \% | 43.91% \%
 \\
& \vca % - baseline
   &  36.6% \% | 36.62% \% 
   & \bf 20.2% \% | 60.32% \% 
   &  21.9% \% | 32.80% \% 
   & \bf 23.3% \% | 28.16% \% 
   &  16.1% \% | 22.60% \% 
   & \bf 4.5% \% | 43.91% \%
 \\
\end{tabular}
}
\caption{Word Error Rates (\%)
of different ASR models
across different subsets within the \linto~audio dataset in Tunisian.
%V0.1 represents the model without any augmentation, including Kaldi's classical augmentations (e.g., speed, volume, and reverberation). V0.2 includes only the standard Kaldi augmentations, while V1 incorporates all augmentations, including classical Kaldi techniques as well as Voice Conversion augmentation.
}
\label{tab:results_asr}
\end{table*}

\subsection{Voice Conversion Augmentation (VCA)}

Several datasets contain single speaker recordings. %, which limited voice diversity and posed challenges for our ASR system.
%To address this issue,
To avoid overfitting certain voices,
we used voice modification to augment single-speaker recordings and increase diversity in an augmented version of the \linto~audio dataset.

% In our study, we compared two prominent voice conversion models: Retrieval-based Voice Conversion (RVC)\footnote{\url{https://github.com/RVC-Project/Retrieval-based-Voice-Conversion-WebUI}} and SoftVC VITS Singing Voice Conversion (So-vits-SVC or SVC)\footnote{\url{https://github.cofromm/voicepaw/so-vits-svc-fork}}. Both models employ a similar encoder-decoder architecture.

In preliminary experiments, we compared several voice modification models
and found that the most realistic results were obtained using SoftVC VITS Voice Conversion~\cite{SoftVC}, %\footnote{\url{https://github.com/voicepaw/so-vits-svc-fork}}, %Singing ~\cite{sovitssvc},
an implementation of Adversarial Training on Soft Speech Units with state-of-the-art methods proposed by~\cite{van_Niekerk_2022,abs-2106-06103}.
We used HuBert~\cite{hsu2021} as a pre-trained encoder,
% to encode speech while disregarding speaker characteristics such as pitch and timbre,
and HiFi-GAN~\cite{kong2020} for the decoder. %, a Generative Adversarial Network (GAN) capable of generating high-fidelity speech with impressive efficiency.
% We amplified these datasets by 5 to 7 times for specific purposes. For instance, voices from ``Abdel Aziz Erwi'', ``Bayari Billionaire'', and others were used in training voice conversion (VC) models. In cases like ``Abdel Aziz Erwi'', we limited augmentation to five times with various speaker models, as this voice was already used to train a VC model, and another VC model shared a similar pitch.
We used a few dozen minutes of speech from 7 speakers, 
including Tunisian story tellers as well as one non native speaker, and trained generative models from those voices.
MASC and data crawled on YouTube (subset ``misc. crawled'')
include a lot of musical clips and especially noisy recordings.
To improve quality, we use a source separation tool called Deezer Spleeter~\cite{spleeter2020} to isolate speech from noise and music. %, resulting in cleaner speech samples. 
% This tool was used for the two training datasets, so each audio file had a noisy and an isolated speech version to increase diversity.
% Applying noise reduction doubled the size of these two datasets. The noise-reduced data durations are part of the train set total under VCA augmentation.
The augmented dataset includes both original and cleaned audio recordings, and VC augmentation was done on cleaned data.

The augmented dataset includes 466 hours of audio,
with details for each subset given in the second column of Table~\ref{tab:audio-data-composition}.

% Our dataset includes various content types, such as noisy recordings and music. In particular, we mention MASC and the misc data Crawled (misc. crawled) from YouTube. This data can hinder clear speech recognition. 
% To improve its quality, we used Deezer Spleeter~\cite{spleeter2020}, a source separation tool, to isolate vocal tracks from musical and noise backgrounds, ensuring clearer speech samples. 
% This tool was applied to both training datasets.
% Additionally, we generated two versions for each audio file --one with noise and one with isolated speech-- to exhibit more diverse audio conditions. 
% For clarity, after applying noise reduction tools, we doubled the data size for both datasets.

\section{Experiments: Tunisian Speech-to-Text}

This section describes preliminary experiments with our training and evaluation datasets, which leads to the release of a first baseline model for Tunisian ASR.

% a detailed overview of the models used for training and fine-tuning our Speech-to-Text (STT) system. To evaluate performance and robustness, we tested the model across multiple datasets and diverse contexts. 
% Our approach initially focused on two main strategies: building a model from scratch using the Kaldi toolkit and fine-tuning a pre-trained Whisper checkpoint.

%\subsection{Baseline results}
First, we tested several large versions of Whisper, as shown in the first rows of Table~\ref{tab:results_asr}.
Even if Whisper can recognize some Arabic,  it fails to transcribe the Tunisian Arabic Dialect, with Word Error Rates (WER) ranging from 50\% to more than 100\% (indicating high insertion rates).

% checkpoints and trained Kaldi models with various configurations.
% These configurations included: (1) a baseline model with no augmentation beyond the inclusion of noise-reduced data, which was necessary due to the presence of musical and highly noisy segments in the dataset; (2) a model employing classical Kaldi augmentations; and (3) a model utilizing both classical and Voice Conversion (VC) augmentations.
% As indicated in Table~\ref{tab:results_asr}, the Whisper models encountered challenges in handling the Tunisian dialect, as the best average result across all tests datasets was only 77\% with the latest large Whisper version.
% After training the Kaldi model from scratch, we addressed two main issues: the difficulty in accurately predicting Tunisian words and the need for real-time transcription, but both of which appear with Whisper.

\begin{figure}[ht]
    \begin{minipage}{0.48\textwidth}
        \centering
        \includegraphics[width=\textwidth]{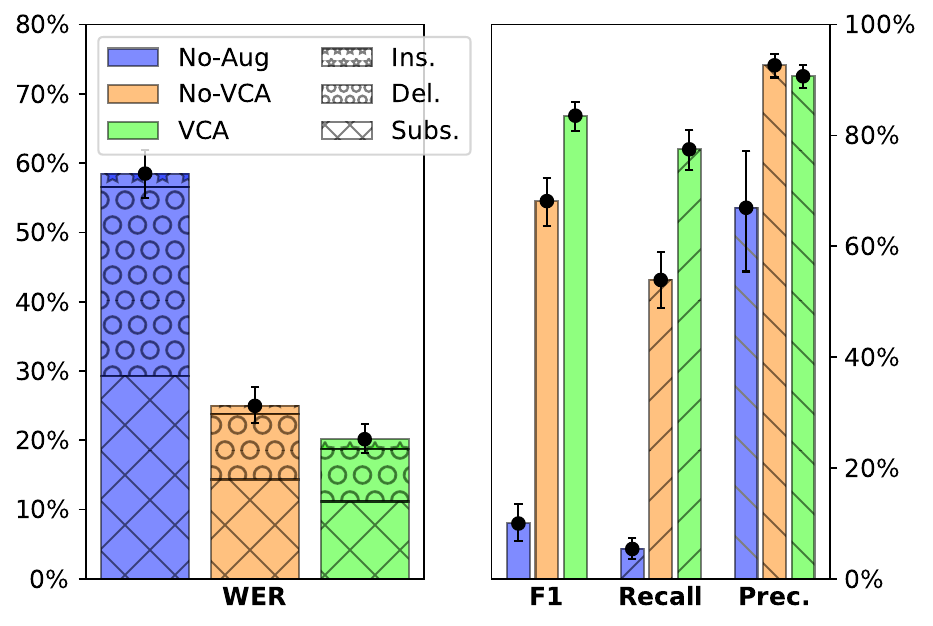}
        \caption{
        Details of results
        across three training conditions
        on TunSwitch Code-Switching (CS) test set.
        {\it left:} Word Error Rates (WER)
        decomposed into insertion (Ins), deletion (Del) and substitution (Subs) rates.
        %: \noaug~(no augmentation), \novca~(Kaldi perturbations without voice conversion), and \vca~(with both voice conversion and Kaldi perturbations).
        {\it right:} F1, Recall and Precision scores on Latin words.
        All 95\% confidence intervals are computed by performing bootstrap resampling.
        }
        \label{fig:bar_graph}
    \end{minipage}
\end{figure}

% \begin{figure}[ht]
%     \begin{minipage}{0.48\textwidth}
%         \centering
%         \includegraphics[width=0.6\textwidth]{figs/ytb_wer.png}
%         \caption{Details of results on YouTube test set.}
%         \label{fig:bar_graph_yt}
%     \end{minipage}
% \end{figure}

%\subsection{Effect of VCA on ASR Model Performance}

% \textcolor{red}{TODO: explain Kaldi~\cite{kaldi} TDNN and some part of the training method (archi, alignment, optimizer), and our augmented Buckwalter transliteration !!! Explain why no phonetization with Arabic (maybe we can find a reference that shows that the gain is little, or ask a reference to the guy of Elya Data, he's the one who explained that)}

To provide a baseline for Tunisian ASR,
we trained  models using the Kaldi open-source toolkit~\cite{kaldi},
%an open-source toolkit for speech recognition. %, used to develop GMM-HMM and DNN-HMM-based ASR systems.
where a triphone GMM-HMM model is trained to align target phonemes with an audio signal. %, which are used as target outputs during TDNN training for accurate phonetic prediction.
The acoustic model consists of a Time Delay Neural Network (TDNN)~\cite{Peddinti2015ATD},
and a Finite-State Transducer (FST) converts sequences of phoneme probabilities into most probable words.
The underlying LM is a word $n$-gram (with $n$ up to 4)
trained on the \linto~textual dataset.
Training details are available under a GPL license.\footnote{
% \textcolor{cyan}{\url{https://github.com/linagora-labs/ASR_train_kaldi_tunisian}}
\url{https://github.com/linagora-labs/ASR_train_kaldi_tunisian}
}

There are no good  phonetic dictionaries for Arabic dialects,
and phoneme-based ASR has failed to outperform  %has never been yet more successfulthan just 
treating each Tunisian Arabic character as a phoneme
\cite{ASR_arabic_phonetic_dict}.
We used Kaldi with the Buckwalter transliteration~\cite{buckwalter}, which we extended to  encode Latin characters while using only ASCII character pairs to represent  Latin and Arabic characters.
% This trick seems to be enough to correctly model code-switching despite ASCII limitations of Kaldi.

% Kaldi only supports ASCII-based phonemes, which poses challenges for Arabic language or dialects since they use Arabic characters, as it's not an ASCII language. Previous work on Modern Standard Arabic (MSA) often used Buckwalter Arabic transliteration which is an ASCII-only scheme to romanize the Arabic words.
% To overcome this challenge, we adopted a method called Buckwalter Augmented, where instead of transliterating all the Arabic text we just transliterate only the phones that exist in the Lexicon. This adaptation enables the model to produce transcriptions directly in Arabic instead of Buckwalter.

Table~\ref{tab:results_asr} compares several training dataset mix conditions for \linto\ models trained on our data:
\begin{itemize}[noitemsep, topsep=0pt]
    \item \noaug: trained on $93$H of original audio data without any augmentation for 20 epochs;
    \item \novca: trained on 4 epochs on $5 \times 93$H  of training data augmented with classical  techniques
    (speed, volume, \ldots). % reverberation, background
    %(speed and volume perturbations\ldots);
    \item \vca's training data also includes VC augmented data, with a total duration of $5 \times 466$H.
\end{itemize}

\vca\ has a notable impact on  code-switching  performance. Figure~\ref{fig:bar_graph} compares on TunSwitch the models trained on the 3 data mixes.
To assess  accuracy for recognition of English and French phrases in a code-switching context,
we give F1, recall and precision scores on Latin words.
% 95\% confidence intervals are computed by performing bootstrap resampling.
We can see that \vca~significantly improves WER, and in particular the recall on Latin words.
These results suggest that \vca~enhances the model's ability to model rarer phenomena, such as English words mixed with Arabic words.
%Despite the improvement on code-switching, it exhibits small regressions on TARIC and YouTube.

% This is illustrated in Figure~\ref{fig:bar_graph_yt}, which shows a 3\% decrease in WER when compared to the \novca model.
% This decline occurs because the VCA model predicts significantly more incorrect words than the \novca model, as evidenced by the substitutions between the two models.

\section{Conclusion}

% In conclusion, this paper demonstrates the effectiveness of our work in building a high-quality and large-scale Tunisian audio and text dataset. Our results indicate that this dataset improves and enhances ASR systems for the Tunisian dialect, offering valuable insights for our ASR system.
% Despite some limitations, such as the limited availability of code-switching resources, this work paves the way for future advancements in AI fields such as multi-modal systems, ASR, NLP, and others.

% Our work contributes to the field of Automatic Speech Recognition (ASR) by releasing public datasets containing both audio and textual data in Tunisian. These datasets not only facilitate the training of ASR models but also provide a valuable resource for their evaluation.
% Our experiments demonstrate that it is feasible to train a reasonably good ASR model using classical training methods on the released dataset.
% We have established a first baseline ASR model for Tunisian Arabic Dialect, showcasing its capabilities, including initial support for code-switching between Tunisian Arabic Dialect and French or English.
% This baseline is the first model trained from scratch exclusively on our open datasets. Future work includes fine-tuning models like Whisper and conformers on the open Arabic spoken data we released and seeing how much further improvement is possible.
Our work advances Automatic Speech Recognition (ASR) by releasing public datasets containing audio and textual data in Tunisian. These datasets can be used to train and evaluate ASR systems.
Through our experiments, we demonstrate the feasibility of training a high-quality ASR model using traditional methods on this dataset. We establish the first baseline ASR model for the Tunisian Arabic Dialect, highlighting its capabilities, including initial support for code-switching with French and English. This baseline is trained exclusively from scratch on our open datasets.
Future work involves fine-tuning models like Whisper and conformers on the open Arabic spoken data we have released.
% \section{Acknowledgments}

% \bigskip
% \noindent Thank you for reading these instructions carefully. We look forward to receiving your electronic files!

\bibliography{biblio}

\end{document}